\documentclass[preprint,12pt, pagewise, left]{elsarticle}

\usepackage{amsmath,amssymb,amsfonts}
\usepackage{graphicx}
\usepackage{hyperref}
\usepackage{algpseudocode}
\usepackage{multirow}
\usepackage[ruled,vlined]{algorithm2e}
\usepackage{adjustbox}
\usepackage{threeparttable}
\usepackage{bm}
\usepackage{textcomp}
\usepackage[misc]{ifsym}
\usepackage{natbib}
\usepackage[table]{xcolor}
\usepackage{lineno} 
\journal{Medical Image Analysis}

\begin{document}
\begin{frontmatter}



\title{Revisiting Medical Image Retrieval via Knowledge Consolidation}


\author[icl]{Yang Nan}
\author[icl]{Huichi Zhou}
\author[icl]{Xiaodan Xing}
\author[au]{Giorgos Papanastasiou}
\author[hkust]{Lei Zhu}
\author[sun]{Zhifan Gao}
\author[cimim,manchester]{Alejandro F Frangi}
\author[icl,nhli,rbh,kcl]{Guang Yang\corref{cor1}}
\cortext[cor1]{send correspondence to g.yang@imperial.ac.uk, y.nan20@imperial.ac.uk}

\affiliation[icl]{organization={Bioengineering Department and Imperial-X, Imperial College},
            city={London},
            country={UK}}
\affiliation[au]{organization={Archimedes Unit, Athena Research Centre},
            city={Athens},
            postcode={15125}, 
            country={Greece}}
\affiliation[hkust]{organization={Hong Kong University of Science and Technology},
            city={Hong Kong},
            country={China}}
\affiliation[sun]{organization={Sun Yat-Sen University},
            city={Guangdong},
            country={China}}
\affiliation[cimim]{organization={CIMIM, Alan Turing Institute},
            city={London},
            country={UK}}
\affiliation[manchester]{organization={NIHR Manchester Biomedical Research Centre, University of Manchester},
            city={Manchester},
            country={UK}}
\affiliation[nhli]{organization={National Heart and Lung Institute, Imperial College London},
            city={London},
            country={UK}}
\affiliation[rbh]{organization={Cardiovascular Research Centre, Royal Brompton Hospital},
            city={London},
            country={UK}}
\affiliation[kcl]{organization={School of Biomedical Engineering and Imaging Sciences},
            city={King's College London},
            country={UK}}

\begin{abstract}
As artificial intelligence and digital medicine increasingly permeate healthcare systems, robust governance frameworks are essential to ensure ethical, secure, and effective implementation. In this context, medical image retrieval becomes a critical component of clinical data management, playing a vital role in decision-making and safeguarding patient information. Existing methods usually learn hash functions using bottleneck features, which fail to produce representative hash codes from blended embeddings. Although contrastive hashing has shown superior performance, current approaches often treat image retrieval as a classification task, using category labels to create positive/negative pairs. Moreover, many methods fail to address the out-of-distribution (OOD) issue when models encounter external OOD queries or adversarial attacks. In this work, we propose a novel method to consolidate knowledge of hierarchical features and optimisation functions. We formulate the knowledge consolidation by introducing Depth-aware Representation Fusion (DaRF) and Structure-aware Contrastive Hashing (SCH). DaRF adaptively integrates shallow and deep representations into blended features, and SCH incorporates image fingerprints to enhance the adaptability of positive/negative pairings. These blended features further facilitate OOD detection and content-based recommendation, contributing to a secure AI-driven healthcare environment. Moreover, we present a content-guided ranking to improve the robustness and reproducibility of retrieval results. Our comprehensive assessments demonstrate that the proposed method could effectively recognise OOD samples and significantly outperform existing approaches in medical image retrieval (p\textless0.05). In particular, our method achieves a 5.6-38.9\% improvement in mean Average Precision on the anatomical radiology dataset.
\end{abstract}

\begin{graphicalabstract}
\includegraphics[width=\textwidth]{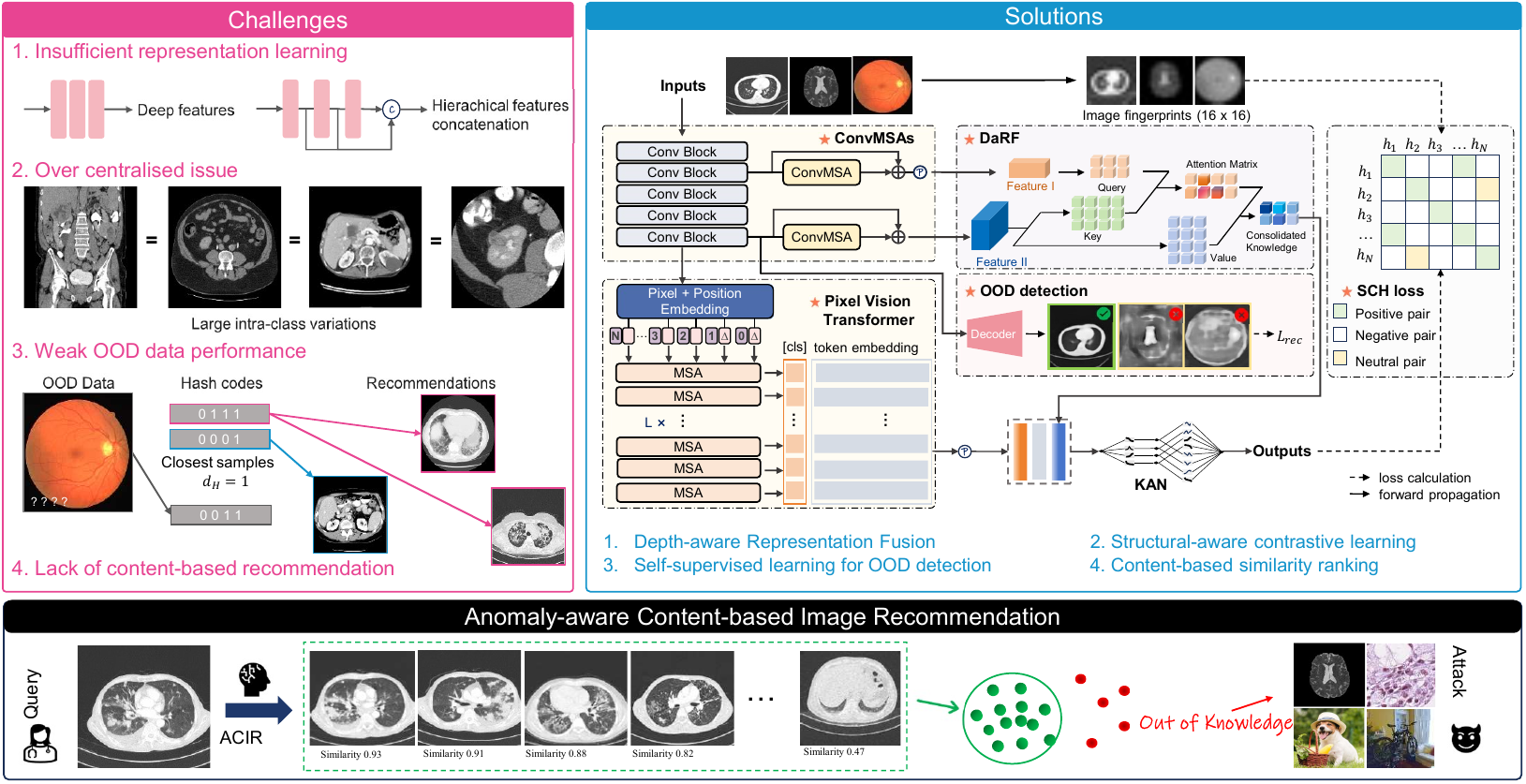}
\end{graphicalabstract}

\begin{highlights}
\item Structure-aware pairing using image fingerprints to address over-centralized issues.
\item A novel model to consolidate hierarchical embeddings for representation learning. 
\item Addressing ill-posed gradient issues introduced by relaxed Hamming distance.
\item A self-supervised OOD detection module by evaluating image reconstruction disparity.
\item Content-guided ranking mechanism for robust and precise retrieval.
\end{highlights}


\end{frontmatter}



\section{Introduction}
\label{sec1}
Improving diagnosis and patient outcomes increasingly depend on medical images, resulting in the accumulation of substantial volumes of imaging data for clinical and research purposes \cite{willemink2020preparing}. Beyond text query systems, image retrieval is crucial for case analysis and clinical instruction, which facilitates quick access to relevant images, supporting efficient diagnosis and disease prognostications for clinicians and AI models. 
\begin{figure}[h]
\centering
\includegraphics[width=0.5\textwidth]{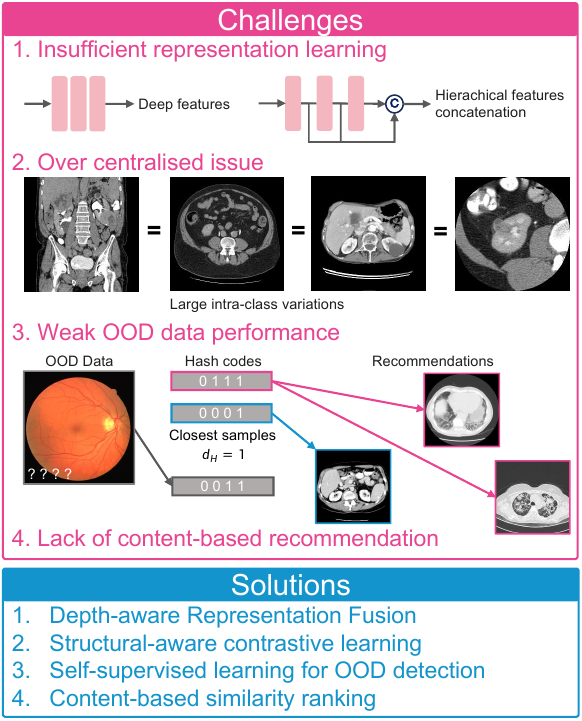}
\caption{Challenges of medical image retrieval, with insufficient representation learning (using deep features or hierarchical concatenation), over-centralised issues (ignoring the intra-class variations), OOD interference (weak capacity against OOD samples), and lack of content-based recommendation.}
\label{fig1}
\end{figure}
However, navigating large-scale image repositories can be challenging, as most current methods struggle to rank images based on content. Interestingly, hashing offers an efficient solution by converting high-dimensional data into binary codes, known as hash codes. These hash codes enable effective similarity searches using the Hamming distance. However, traditional hashing techniques often produce imprecise results due to their limited feature extraction capabilities. To bridge these gaps, researchers have proposed deep-learning approaches \cite{li2018large} for image retrieval and significant improvements have been witnessed. In particular, existing deep hashing approaches could be divided into two types: dynamic hashing (including doublet and triplet hashing) and fixed hashing. Dynamic hashing utilizes deep contrastive learning to encourage positive pairs to be close in latent space while pushing negative pairs apart \cite{yang2023semantic}. Fixed hashing treats the retrieval task as an alternative classification task, by training the network to learn a predefined hash code for each category \cite{csq,dpn}. 

In spite of continued efforts, numerous challenges remain in medical image retrieval. (1) \textbf{Inadequate representation learning.} The majority of methods produce hash codes utilising features of the final bottleneck or simply concatenating hierarchical embeddings. These techniques fail to adjust to both deep and shallow features, resulting in inadequate representation learning. (2) \textbf{Over-centralisation.} The goal of image retrieval is to locate an image of the same category as the query image, typically achieved through contrastive training on pairs of positive and negative images. However, substantial visual discrepancies between samples in the same category can hinder model convergence (Fig. 1). The intraclass variations complicate the model's ability to learn a general representation applicable to all samples in the category. Furthermore, most methods utilise the relaxed Hamming distance to measure the distance, which causes an unstable training process because of abrupt probability functions. (3) \textbf{Weak out-of-distribution capability.} Existing methods show poor performance on out-of-distribution (OOD) data, as the model tends to assign an OOD sample to a known category, reflecting weak generalisability and robustness. (4) \textbf{Deficiency in content-based recommendation.} Most current research does not generate ranking results based on content similarity.

To address these challenges, this study introduces the \textbf{A}nomaly-aware \textbf{C}ontent-based \textbf{I}mage \textbf{R}ecommendation (\textbf{ACIR}), aiming to achieve high-performance retrieval and accurately rank relevant images according to the content of the image. (1) To capture both low-level and high-level representations, we propose a Depth-aware Representation Fusion (DaRF) module to ensemble multi-level features, followed by a pixel vision Transformer (PiT) to globally refine deep features from Convstem. (2) For the over-centralised issue, we introduced a structure-aware pairing mechanism using image fingerprints into the contrastive hashing. This helps ACIR to assign neutral labels to positive image pairs with large intraclass variations. In addition, we adopted the Pearson coefficient to measure the sample distance in latent space, alleviating the unstable training process caused by sharp probability functions. (3) To address OOD hallucination, a self-supervised decoder was introduced to recognise OOD samples. (4) Content-based classification was achieved by calculating the average similarity between hierarchical consolidated embeddings. To thoroughly evaluate the full capacity of the proposed method in medical datasets, we performed experiments on multianatomy datasets comprising RadioImageNet-CT (29903 greyscale images with 16 classes) and the Breast Cancer Semantic Segmentation (BCSS) dataset (18,678 RGB images with 6 classes). The main contributions can be summarised as follows:
\begin{itemize}
    \item We introduced a structure-aware pairing mechanism utilizing image fingerprints to assign neutral labels, addressing the over-centralization problem.
    \item We proposed a novel architecture integrating ConvMSAs, PiT, and a depth-aware fusion module to enhance hierarchical embedding consolidation for efficient representation learning.
    \item To mitigate ill-posed gradient issues, we employed the Pearson correlation coefficient as an alternative to the relaxed Hamming distance for estimating sample distances.
    \item A self-supervised OOD detection module was developed, leveraging the disparity in image reconstruction outputs for robust reliable retrieval.
    \item We proposed a content-guided ranking mechanism to enable robust and precise retrieval.
\end{itemize}

\section{Related Works}
\noindent Given the query image, a deep hashing model $\phi(\Theta)$ aims to find correlated images by mapping raw images $X = \{x_1, x_2, \ldots, x_N\}$ into K-bit binary hash vectors $H = \{h_1, h_2, \ldots, h_N\}$, with $\phi: X \rightarrow H, h_N \in [-1, 1]^K$, followed by signing $H$ into binary hash codes $B \in \{0, 1\}^{N \times K}$
\begin{equation}
    B = \text{sgn}(f(\phi(X; \Theta)))
\end{equation}
where $\Theta$ indicates trainable parameters of the network, $f$ is an activation function that can be linear or non-linear, $\text{sgn}$ is the function that returns the signs of the elements of the input. The optimization functions for deep hashing typically consist of two key components, with a distance loss $L_D$ to maximize(minimize) negative(positive) pairs and a quantisation loss $L_Q$ to quantise hash vectors to binary hash codes. The `distance' is mainly estimated through the relaxed Hamming distance $d_{RH} = \mathcal{P}(d_H)$, using probability functions $\mathcal{P}$ to map the Hamming distance to $[0, 1]$. These probability functions mainly include the sigmoid $\sigma$  (shifted as $\mathcal{P}(d_H) = \sigma (d_H - \frac{K}{2})$), Gaussian \cite{tu2021weighted}, and tanh \cite{jiang2018asymmetric}. The quantisation loss aims to quantise hash vectors to binary hash code, which is commonly presented as
\begin{equation}
L_Q = \| \mathbf{1} - |h|\|_1,
\end{equation}
where $\mathbf{1}$ is an all-one vector of length $K$ \cite{yan2020deep}. Furthermore, some researchers adopted sigmoid or tanh activation functions to restrict the output of networks to $[-1, 1]$.

Existing studies mainly focus on several aspects: 1) developing novel networks. For instance, ATH \cite{ATH} incorporated a spatial-attention module and introduced a triplet cross-entropy loss to enhance class separability. FIRe \cite{FIRe} used iterative attention modules to learn localized and discriminative image patterns; 2) exploring novel learning mechanisms. For example, MTH \cite{MTH} employed triplet hashing instead of conventional double-paired hashing to capture discriminative areas and hierarchical similarity. CSQ and DPN \cite{csq, dpn} predefined hash codes to maximize the distance between hash codes of different categories. CenterHash embeds multi-modal data into a unified Hamming space while effectively learning discriminative hash codes to address the challenges posed by imbalanced multi-modal images \cite{yang2020deep}; and 3) proposing advanced optimization functions for efficient representation learning \cite{zheng2020deep,chen2019deep, qsmih}. For instance, Cao et al. employed Cauchy \cite{dch} to generate compact and concentrated binary hash codes. DBDH \cite{dbdh} reduced quantization error by avoiding continuous relaxation, directly generating binary codes through discrete gradient propagation. Despite significant efforts, most studies have overlooked the inherent intra-class variation, placing excessive emphasis on learning discriminative feature representations. Furthermore, none of the existing models have demonstrated the capability to identify OOD samples, resulting in potential risks in practical applications.

\section{Methods}
\noindent An overview of the proposed ACIR is demonstrated in Fig. 3. ACIR mainly comprises four parts, with a hybrid encoder ConvPiT and a depth-aware representation fusion module for learning high-efficient representations, structure-aware contrastive learning for addressing the over-centralised issue, a reconstruction module for OOD detection, and content-guided ranking for ranked retrieval.
\vspace{-0.3cm}
\subsection{Problem Definition}
\noindent Given a set of images $\mathbf{X}$ and their labels $L = \{l_1, l_2, \ldots, l_N\}$, the semantic similarity $s_{i,j}$ between image pairs $x_i$ and $x_j$ is defined by their class label
\begin{equation}
s_{i,j} = \begin{cases}
1, & l_i = l_j \\
0, & l_i \neq l_j.
\end{cases}
\end{equation}
With the similarity matrix $S = \{s_{0,0}, s_{0,1}, \ldots, s_{N,N}\}$ of $N$ paired samples, the hash model $\phi(\Theta)$ can be trained by maximizing a posterior estimation of binary embedding $H = \{h_1, h_2, \ldots, h_K\}$
\begin{equation}
\hat{\Theta}_{\text{MAP}} = -\arg \min_{\theta} (\log P(S|H) + \log P(H)),
\label{hash_loss}
\end{equation}
where $\log P(H)$ is the prior probability of $H$, $\log P(S|H)$ is the log-likelihood estimation. The conditional probability of the similarity $s_{i,j}$ given two hash vectors $h_i, h_j$ is calculated by
\begin{equation}
P(s_{i,j} | h_i, h_j) = \begin{cases} 
\mathcal{P} \left( \mathcal{D} (h_i, h_j) \right), & s_{i,j} = 1 \\ 
1 - \mathcal{P} \left( \mathcal{D} (h_i, h_j) \right), & s_{i,j} = 0 
\end{cases}
\end{equation}
where $\mathcal{D}$ refers to the distance measurement that evaluates how `close' $h_i, h_j$ is, and $\mathcal{P}$ is the probability function that converts the distance (e.g., hamming distance) to probabilities.

\begin{figure}
    \centering
    \includegraphics[width=0.95\linewidth]{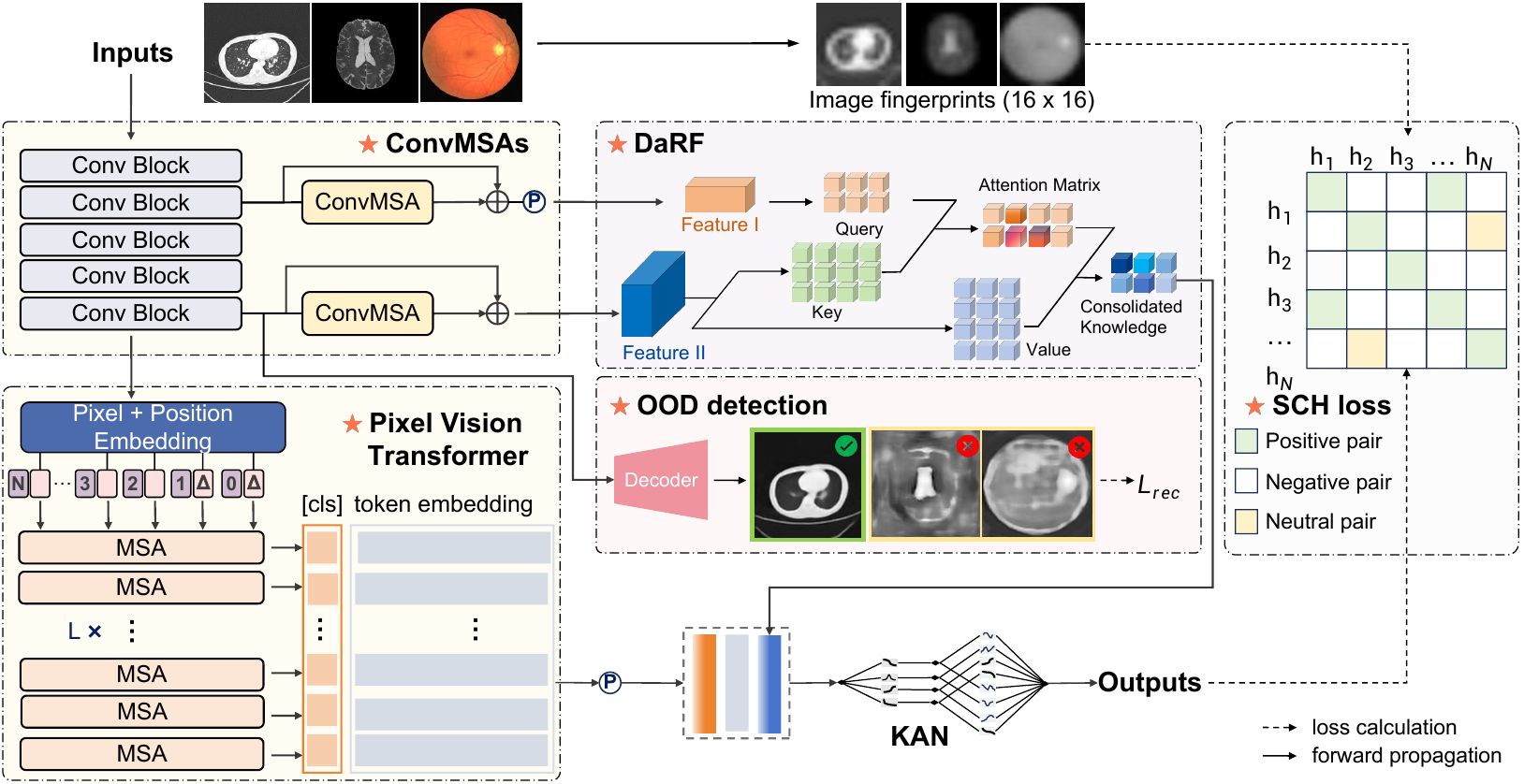}
    \caption{Overview of the proposed method. ACIR leverages images and their corresponding low-resolution fingerprints for structure-aware contrastive hashing (SCH). The hash embeddings \textbf{H} are obtained by Convolutional Multi-head self-attention layers (ConvMSAs), Depth-aware Fusion (DaRF) and Kolmogorov-Arnold Network (KAN). Additionally, a self-supervised module is integrated to enhance Out-of-Distribution (OOD) detection capabilities.}
    \label{fig:overview}
    \vspace{-0.3cm}
\end{figure}

\vspace{-0.3cm}
\subsection{Hierarchical Knowledge Consolidation}
\noindent Effective models for image retrieval are required to map raw data into representative embeddings, ensuring that samples belonging to the same category are aptly clustered around their respective hash centroids. This section presents how ACIR extracts and consolidates hierarchical knowledge, including using PiT to globally refine high-level features, and DaRF to adaptively integrate multilevel representations.

Although Vision Transformers (ViTs) are proficient in capturing global features, they demand significantly more training data and computational costs compared to Convolutional Neural Networks (ConvNets) \cite{vit}. Conversely, pure ConvNets struggle to capture global feature representations due to their inherent inductive biases of locality and translational invariance. To extract high-quality hierarchical information from input images, the ACIR architecture employs a hybrid approach: it uses a ConvNet stem to extract both shallow and deep features (Fig. \ref{fig:overview}), and then a pixel transformer (PiT) globally refines these features. Research indicates that combining ConvNets with transformers can counteract the Transformer's locality limitations (resulting from tokenising input images) and perform better than standard ViTs \cite{heo2021rethinking, FIRe}. However, this advantage diminishes for larger ViT models as they employ smaller token sizes to address the locality issue in tokenisation. Although reducing token sizes can help address locality problems, it considerably increases the computational burden. Understanding the importance of cross-token interactions for representation learning, we stress that calculating the interactions between different tokens and assessing their influence via the self-attention mechanism is essential. Therefore, ACIR treats each pixel ($1\times 1$) as a token with pixel-wise positional embedding (PE) for comprehensive cross-token calculations.\begin{equation}
    X = [\mathrm{cls}, p_0, p_1, ..., p_n] + \mathrm{PE}
\end{equation}
where \text{cls} represents the class token and $p_n$ denotes the count of flattened pixels (involving both height and width axes) within the feature map. Once these pixel embeddings are obtained, the $L$ MSA layers further refine the deep features to achieve globally refined embeddings. These globally refined embeddings are derived from all pixel tokens (as shown by blue squares in Fig. \ref{fig:overview}) and the [cls] token (represented by orange squares) symbolizes the global representations. 

To further address small inter-class variations, ACIR incorporates low-level and high-level features from the Convstem, refining them through convolutional multi-head self-attentions (ConvMSAs) and depth-aware fusion for knowledge consolidation. ConvMSAs compute self-attention within convolutional receptive fields by unfolding the feature maps. The features processed by ConvMSA are enriched with local-global context and enhanced spatial relationship representation. In vanilla MSA, the Query (Q), Key (K), and Value (V) are with the size of $N \times C$ (where N is the sequence length and C is the embedding dim). In CovnMSA, given the input \( x_{in} \in \mathbb{R}^{B \times C \times H \times W} \), the window size $(p_h, p_w)$, embeddings are first rearranged from $B\times C \times (n_h p_h) \times (n_w p_w)$  to $(B n_h n_w)\times C \times p_h \times p_w$, where $H=n_h p_h$ and $W=n_w p_w$. Then $Q'$, $K'$ and $V'$ are obtained through $1\times1$ conv layers 
\begin{equation}
    Q', K', V' = \mathcal{C}_{1\times1}(C, D_h*N_h)(x_{in}),
\end{equation}
where $N_h$ refers to the number of heads, $D_h$ indicates the head dims, and $C=D_h\times N_h$. \bm{$\mathcal{C}_{1\times1}(C_{in}, C_{out})(\cdot)$} represents a $1\times1$ Conv layer taking $x_{in}$ (with the shape of $B n_h n_w\times C_{in}\times p_h \times p_w$) as input and outputs a ($B n_h n_w\times C_{out}\times p_h \times p_w$) tensor. The $Q'$, $K'$ and $V'$ are then reshaped and rearranged to facilitate the attention mechanism, transforming into dimensions suitable for multihead attention (from $B\times D_h N_h \times p_h \times p_w$ to $B\times N_h \times p_h p_w \times D_h$). Subsequently, the attention is calculated by applying softmax to the scaled dot product between the $Q'$ and $K'$.
\begin{equation}
    \mathrm{Attention}(Q',K',V')=\mathrm{Softmax}(\frac{Q'K'^T}{\sqrt{D_{h}}})V'.
\end{equation}
After the self-attention, the output is reshaped to the raw input size ($B\times C \times (n_h p_h) \times (n_w p_w)$).

In addition to ConvMSA, DaRF aims to adaptively consolidate features from different hierarchical levels to generate comprehensive feature embeddings. The shallow features $E_s$ capture primarily low-level semantic meaning but are rich in textual information, while the deep features $E_d$ carry more semantic meaning but have fewer textual details. These pieces of knowledge were first gained from the shallow (2nd) and deep (last) conv blocks, followed by ConvMSAs for local self-attention. After ConvMSAs, the outputs are aligned to the same shape by adaptive pooling and fed into DaRF for feature fusion.  Using low-level knowledge to update high-level ones, we leverage the semantic guidance from the shallow features to identify and focus on the corresponding deep representations.
\begin{equation}
    R_F=\mathrm{Softmax}(\frac{\mathcal{C}_{1\times1}(\mathcal{I}(E_d))\mathcal{C}_{1\times1}(\mathcal{A}(E_{s}))^T}{\sqrt{D_{h}}})E_s,
\end{equation}
which could help the model to generate more distinguishable embeddings. 

After the DaRF, the fused 2D representations $R_F$ were projected to the same channel as the input $x_{in}$ and were further compressed to 1D vectors by global average pooling, then concatenated with global and refined representations tokens from MSA layers. Subsequently, we used two Kolmogorov-Arnold layers (KAN) \cite{liu2024kan} as an alternative to multilayer perceptions, which have learnable activation functions on edges to further consolidate multilevel knowledge. Given the input $x_p$, a KAN layer with $C_{in}$-dimensional inputs and $C_{out}$-dimensional outputs is defined as
\begin{equation}
    \mathrm{KAN}(x)=\{\phi_{p,q}\},\quad p=1,2,...,C_{in},\quad q=1,2,...,C_{out},
\end{equation}
where 
\begin{equation}
   \phi(x)=w_bb(x)+w_s\mathrm{spline}(x), 
\end{equation}
where $b(x)$ and $\mathrm{spline}(x)$ are the basis function and spline function, respectively, $w_b$ and $w_s$ refer to trainable weights. In this study, we applied one KAN layer to convert $x_p$ into hash vectors $h\in \mathbb{R}^\mathrm{bit}$.

\vspace{-0.3cm}
\subsection{Structural-aware Contrastive Hashing}
\noindent With the consolidated embeddings, the following step is to learn a hash function to convert these embeddings into hash codes. Based on Eq. \ref{hash_loss}, most studies quantified the similarity of sample pairs by relaxing the Hamming distance with probability functions (e.g., using sigmoid, tanh, and Gaussian to rescale the distance to [0,1]). However, these functions present a sharp (flat) shape when the Hamming distance is considerably small (large), leading to ill-posed gradient issues. Moreover, these relaxed Hamming distances only assess the magnitude between samples, which aggravates the over-centralisation of learned hash vectors. Although some researchers applied cosine similarity to assess the direction of hash vectors \cite{jang2020generalized}, the magnitude assessment was ignored.

To bridge this gap, we aim to assess the 'distance' between different samples by considering both the magnitude and direction of hash embeddings. As a result, the Pearson coefficient is introduced for measuring both magnitude and direction similarity, with
\begin{equation}
    S(\mathbf{x}, \mathbf{y}) = \frac{\sum (x_{n,k} - \bar{x}_k)(y_{n,k} - \bar{y}_k)}{\|x_{n,k} - \bar{x}_k\|_2 \|y_{n,k} - \bar{y}_k\|_2}.
\end{equation}
Therefore, the contrastive loss $L_C$ is calculated by
\begin{equation}
    L_C = -s_{i,j}\log{S(h_i, h_j)}-(1-s_{i,j})\log(1-S(h_i, h_j)),
    \label{lc}
\end{equation}
Meanwhile, we formulate the quantisation loss as
\begin{equation}
L_Q = \log (\lambda - S(h_i|\mathbf{1}, 1)),
\end{equation}
where \(\mathbf{1}\) is an all-ones vector which has the same length as the hash embeddings \(h_i\). $\lambda$ is set to 2 to avoid numerical errors.

To alleviate over-centralisation, we weighted the distance of samples within the same category that exhibit considerable visual differences (Fig. 1). Unlike natural images, medical images retain essential structural information even at low resolution, which is crucial for content-based retrieval. Thus, we downsampled the original images to create low-resolution fingerprints, which were used to calculate structural consistency (Fig. \ref{fig:overview}). Based on this prior knowledge, we implemented a structure-aware pairing mechanism, categorising samples into positive pairs (the same class, similar appearance), neutral pairs (same class, visual discrepancy), and negative pairs (different classes). By adding the neutral pairs, the structure-aware contrastive loss is now presented as 
\begin{equation}
    L_{SC} = -\mathbf{H}*s_{i,j}\log{S(h_i, h_j)}-e^\mathbf{H}(1-s_{i,j})\log(1-S(h_i, h_j)),
\end{equation}
where $\mathbf{H}$ indicates the Hermite matrix that illustrates the structural consistency. By introducing the constraint $e^\mathbf{H}$, the model enhances penalties for samples from different categories that have small inter-class variations. This ensures that these subtle differences are magnified, aiming to better separate the categories.  Meanwhile, the Hermite matrix $\mathbf{H}$ also alleviates penalties for neutral pairs (e.g., CT images of the same class but from different planes) to increase the tolerance of intra-class variation. 

Besides the over-centralised issue, another challenge is the class imbalance, leading to a bias towards preference retrieval. Although solutions for data imbalance have been presented in \cite{hashnet,dch}, these works mainly focused on the imbalance between positive and negative pairs, thus ignoring the variation for negative pairs. In other words, all negative pairs are considered the same class during contrastive learning. To address this issue, we formulate a class weight matrix that can be directly integrated into contrastive learning. Given \(N\) samples that belong to the categories \(C\), the class weight matrix $W_M$ is denoted as $W_M = \mathbf{w}^T \mathbf{w}$, where \(\mathbf{w} = \{w_1, w_2, \ldots, w_C\}\) is the class weights, with 
\begin{equation}
w_i = \frac{N}{N_i * C},
\end{equation}
where \(N_i\) is the number of class \(i\) samples.
Therefore, the weighted structure-aware contrastive loss can be presented as 
\begin{equation}
    L_{WSC}=W_ML_{SC} + \alpha L_Q
\label{loss}
\end{equation}
where $\alpha$ was empirically set to 0.5 in our experiments.

\vspace{-0.3cm}
\subsection{Out-of-distribution detection}
Studies have shown that the reconstruction model struggles to rebuild OOD samples and produces higher reconstruction errors \cite{li2023rethinking}. In this study, we designed a light decoder to reconstruct the original input using deep representations from the ConvNet stem. Additionally, skip connections were removed to force the decoder to reconstruct inputs using only the given feature representations. Specifically, the decoder of ACIR includes a set of Conv blocks combined with Conv layers, Instance Normalisation (IN), LeakyReLU and Upsampling layers, optimized by $l1$ loss. To reduce the computational cost, the Upsampling scales were set to 2, 4, and 4, respectively.

To evaluate the reconstruction error of grey images, the multi-scale structure similarity ($\mathcal{M}_{sim}$) \cite{wang2003multiscale} is adopted. Given the exponential weights $\alpha_M$, $\beta_j$, and $\gamma_j$, $\mathcal{M}_{sim}$ estimates the similarities between $x$ and $y$ at M scale
\begin{equation}
\mathcal{M}_{sim} = [l_M(x,y)]^{\alpha_M} \prod_{j=1}^{M} [c_j(x,y)]^{\beta_j}[s_j(x,y)]^{\gamma_j}.
\end{equation}
where $M$ is the scale factor, $l_M$, $c_j$, and $s_j$ are the luminance, contrast and structure comparison measures. Therefore, OOD data can be recognized by calculating the $\mathcal{M}_{sim}$ between the input images and their reconstructed ones. In particular, given the $i$-th raw input image $X_i$ and the $i$-th reconstructed images $R_i$, the threshold for OOD data detection is set by
\begin{equation}
\tau = \frac{1}{N} \sum_{i=1}^{N} \mathcal{D}(X_i, R_i) \pm 3 \ast \delta(\mathcal{D}(X, R)),
\label{tao}
\end{equation}
where $\mathcal{D}$ is the distance measurement such as $\mathcal{M}_{sim}$, $\delta$ is the standard deviation. For $\mathcal{D}$, we utilized $1-\mathcal{M}_{sim}$ (Eq. \ref{tao}) for RGB images and the $L_1$ norm for grayscale images, respectively.

\vspace{-0.3cm}
\subsection{Content-based similarity ranking}
Given a query image $X_Q$, ACIR first identifies all potential candidates from the database $X_D$ by assessing the Hamming distance within the hamming balls. Subsequently, the residual ($L_1$ or $1-\mathcal{M}_{sim}$) between $X_Q$ and its reconstruction $R_Q$ is evaluated. $X_Q$ is identified as OOD data if its reconstruction residual $E_R$ surpasses the OOD threshold $\tau$.

\begin{algorithm}[!h]
    \caption{Content-based Ranked Recommendation}
    \label{alg:ranked_retrieval}
    \KwIn{Query image $X_Q$, database $X_D = \{ X_1, X_2, \dots, X_n \}$, model ACIR $\mathcal{T_{\theta}}$}
    \KwOut{Recommendations $X_T = \{ X'_1, X'_2, \dots, X'_m \} \subset X_D$}
    \BlankLine
    Step 1: Obtain hash codes, embeddings, and reconstructions of $X_Q$ and $X_D$ through $\mathcal{T_{\theta}}$. $H_Q, E_Q, R_Q=\mathcal{T_{\theta}(X_Q)}; H_D, E_D, \_\mathcal{=T_{\theta}(X_D)}$.
    
    Step 2: Calculate the reconstruction error $E_R=\mathcal{D(X_Q, R_Q)}$

    Step 3: \If{$E_R$ \textgreater $\tau$}{$X_Q \in OOD$, break}
    
    Step 4: Calculate the hamming distance $D_h=(H_Q, H_D)$ and the content similarity list $L_{sim}=PCC(E_Q, E_D)$. 
    
    Step 5: Obtain top k candidates based on sorting content similarity and hamming distance $lexsort(-L_{sim}, D_h)$.

\end{algorithm}

However, relying on the Hamming distance to determine the top-\(K\) samples may raise issues when a large number of samples share the same (minimum) Hamming distance to \(X_Q\). For instance, one may find 1{,}000 candidates with a distance of \(0\), while only top-200 samples are requested. A purely Hamming-based ordering cannot further distinguish among these equally distant samples. To address this issue, we used content similarity as an additional ranking criterion. Specifically, we extracted multilevel feature representations (from encoder blocks) and calculated the average Pearson correlation coefficient (PCC) to capture deeper content-based relationships between samples. When multiple candidates are tied at the same Hamming distance, this content similarity score provides a more fine-grained differentiation.

\section{Experiments and Results}
\subsection{Datasets and Training Details}
\noindent We conducted experiments on two publicly available datasets. 

\noindent\textbf{BPS:} BPS includes 18703 images collected from \cite{amgad2019structured} and \cite{graham2021lizard}, with 6 categories of tumour (n=6659), stroma (n=4052), inflammatory (n=2798), necrosis (n=1876), fat (n=1674), and gland (n=1619). Specifically, images were split into training (n=13092), valid (n=3740), and test (n=1871) sets.

\noindent\textbf{RadIN-CT:} RadIN-CT, based on RadioImageNet \cite{mei2022radimagenet}, includes multiple anatomical cases with sixteen clinical patterns: normal abdomen (n=2158), airspace opacity (n=1987), bladder pathology (n=1761), bowel abnormality (n=2170), bronchiectasis (n=2025), interstitial lung disease (n=2278), liver lesions (n=2182), normal lung (n=2028), nodule (n=2029), osseous neoplasm (n=1584), ovarian pathology (n=1471), pancreatic lesion (n=2165), prostate lesion (n=948), renal lesion (n=2200), splenic lesion (n=653), and uterine pathology (n=2194). The dataset comprises mostly axial plane images, with some noisy, magnified, and coronal view samples. The 29903 images are split into training (n=20932), validation (n=5980), and test (n=2991) sets.

\noindent\textbf{Training Strategy:} 
All trainable parameters were initialized with He-normal initialisation. During the training, random flips (up, down, left, right) and resized crops (scale $\in [0.8, 1.0]$) were used. Additionally, brightness, contrast, saturation, and hue adjustments were randomly applied to the BPS dataset. Models were trained on two NVIDIA RTX 3090 Ti GPUs for 400 epochs using the AdamW optimizer with an initial learning rate of 3e-4 and a cosine learning rate scheduler ($\mathrm{lr}{min}$=2e-6, $\mathrm{lr}{wp}$=1e-4, $\mathrm{epoch}_{wp}$=5). All comparisons were trained following their original papers (with ImageNet pretrained weights). 

\vspace{-0.3cm}
\subsection{Experimental Settings and Evaluation Metrics}
\noindent\textbf{Comparison studies}. We conducted comparison studies on three datasets by assessing central similarity quantisation (CSQ) \cite{csq}, deep polarised network (DPN) \cite{dpn}, deep Cauchy hashing (DCH) \cite{dch}, deep balanced discrete hashing (DBDH) \cite{dbdh}, quadratic spherical mutual information hashing (QSMIH) \cite{qsmih}, ViT-LLaMa \cite{vitllama}, Attention-based Triplet Hashing (ATH) \cite{ATH}, Feature Integration-based Retrieval (FIRe) \cite{FIRe}, and Multi-scale triplet hashing (MTH) \cite{MTH}. All these comparisons were published in high-impact journals/conferences and most of them (except MTH) are open-source available.

\noindent\textbf{Ablation studies} 
To better understand the effectiveness of different modules, ablation studies were conducted on the RadIN-CT dataset. We set up three baselines, with ConvNet (ResNet50), ViT (ViT-base), and our hybrid model ResNet50+PiT. These baselines were first trained with Cauchy cross-entropy loss \cite{dch} and subsequently enhanced with $L_{WSC}$, DaRF, and KAN for ablation studies. Additionally, the hyper-parameter selection of $\alpha$ in Eq. (\ref{loss}) was investigated, with its values ranging from 0 to 1 in increments of 0.25.
All these models were trained following the same training strategies (with pretrained weights) for a fair comparison. 

\noindent\textbf{Out-of-distribution assessment}
To assess the capacity of different models against OOD samples, top-4 approaches were evaluated on an additional OOD dataset. We randomly selected 8000 samples (500 samples for each category) from validation and test sets (of RadIN-CT) combined with an additional out-of-distribution class 'OOD'(n=500). In particular, these 500 OOD images were randomly selected from the retina \cite{hoover2000locating}, brain MRI \cite{calabrese2022university}, BPS, cell imaging \cite{sypetkowski2023rxrx1}, and VOC2012 \cite{everingham2015pascal} datasets. The hash centres were identified by averaging their corresponding hash codes. OOD samples were identified when the corresponding hash codes were located outside of the Hamming balls ($r = \frac{bit}{4} + 1$), relative to their hash centre. In addition to the above criteria, ACIR further recognises OOD samples based on the threshold $\tau$ in Eq. \ref{tao}.

\noindent\textbf{Evaluation metrics}
The performance was evaluated through mean Average Precision(mAP) and macro Average Precision(maAP). True positive predictions are counted when (1) the retrieved samples belong to the same class as the query image, and (2) the hash code of the query image matches or is close to (within the Hamming balls) that of the retrieved samples. Additionally, samples within the Hamming ball are regarded as true positive samples. During the evaluation, we mainly assessed the top 100 retrieved samples to report the mAP. The significance is given by conducting the Wilcoxon signed rank test between the evaluation results derived using ACIR and the comparison methods, with P \textless 0.05 indicating significant differences between the two paired methods.

\vspace{-0.2cm}
\subsection{Experimental Results}
\noindent\textbf{Comparison experiments on RadIN-CT}
The comparison results presented in Table I for the RadIN-CTL dataset highlight the superior performance of the proposed method (ACIR). 

\begin{table}[!ht]
\vspace{-0.3cm}
\centering
\caption{Comparison Studies on BPS and RadIN-CTL}
\begin{adjustbox}{width=\textwidth}
\begin{tabular}{c|cccc|cccc}
\hline
\multirow{2}{*}{Model} & \multicolumn{4}{c|}{RadIN-CT (C=16)} & \multicolumn{4}{c}{BPS (C=6)} \\
\cline{2-9}
& 8 bits & 16 bits & 64 bits & 128 bits & 8 bits & 16 bits & 64 bits & 128 bits \\
\hline
CSQ & 0.660(0.641)$^\dagger$ & 0.657(0.628)$^\dagger$ & 0.650(0.632)$^\dagger$ & 0.655(0.635)$^\dagger$ & \textbf{0.970(0.967)} & 0.960(0.956)$^\dagger$ & 0.963(0.953)$^\dagger$ & 0.954(0.952)$^\dagger$ \\
DPN & 0.573(0.528)$^\dagger$ & 0.656(0.630)$^\dagger$ & 0.660(0.636)$^\dagger$ & 0.658(0.628)$^\dagger$ & 0.951(0.947)$^\dagger$ & 0.964(0.953)$^\dagger$ & 0.961(0.952)$^\dagger$ & 0.968(0.969)$^\dagger$ \\
DBDH & 0.508(0.470)$^\dagger$ & 0.580(0.551)$^\dagger$ & 0.556(0.512)$^\dagger$ & 0.569(0.544)$^\dagger$ & 0.756(0.718)$^\dagger$ & 0.738(0.670)$^\dagger$ & 0.745(0.695)$^\dagger$ & 0.738(0.692)$^\dagger$ \\
DCH & 0.608(0.577)$^\dagger$ & 0.612(0.582)$^\dagger$ & 0.602(0.568)$^\dagger$ & 0.589(0.553)$^\dagger$ & 0.949(0.939)$^\dagger$ & 0.952(0.940)$^\dagger$ & 0.947(0.939)$^\dagger$ & 0.931(0.920)$^\dagger$ \\
DSDH & 0.449(0.417)$^\dagger$ & 0.464(0.445)$^\dagger$ & 0.344(0.325)$^\dagger$ & 0.355(0.333)$^\dagger$ & 0.934(0.926)$^\dagger$ & 0.933(0.927)$^\dagger$ & 0.930(0.929)$^\dagger$ & 0.934(0.933)$^\dagger$ \\
QSMIH & 0.492(0.463)$^\dagger$ & 0.601(0.568)$^\dagger$ & 0.619(0.589)$^\dagger$ & 0.627(0.592)$^\dagger$ & 0.933(0.931)$^\dagger$ & 0.939(0.940)$^\dagger$ & 0.945(0.946)$^\dagger$ & 0.947(0.944)$^\dagger$ \\
ViT-LLaMa & 0.613(0.582)$^\dagger$ & 0.598(0.579)$^\dagger$ & 0.627(0.603)$^\dagger$ & 0.644(0.619)$^\dagger$ & 0.965(0.960)$^\dagger$ & 0.963(0.960)$^\dagger$ & 0.965(0.962)$^\dagger$ & 0.971(0.969)$^\dagger$ \\
FIRe & 0.613(0.577)$^\dagger$ & 0.634(0.607)$^\dagger$ & 0.658(0.628)$^\dagger$ & 0.651(0.613)$^\dagger$ & 0.946(0.934)$^\dagger$ & 0.958(0.961)$^\dagger$ & 0.956(0.941)$^\dagger$ & 0.948(0.940)$^\dagger$ \\
ATH & 0.327(0.287)$^\dagger$ & 0.352(0.316)$^\dagger$ & 0.352(0.314)$^\dagger$ & 0.341(0.302)$^\dagger$ & 0.466(0.318)$^\dagger$ & 0.459(0.312)$^\dagger$ & 0.474(0.332)$^\dagger$ & 0.449(0.320)$^\dagger$ \\
MTH* & 0.494(0.473)$^\dagger$ & 0.514(0.489)$^\dagger$ & 0.560(0.536)$^\dagger$ & 0.560(0.531)$^\dagger$ & 0.967(0.962) & 0.953(0.955)$^\dagger$ & 0.962(0.963)$^\dagger$ & 0.967(0.965)$^\dagger$ \\
ACIR(Ours) & \textbf{0.683(0.661)} & \textbf{0.701(0.680)} & \textbf{0.706(0.690)} & \textbf{0.716(0.701)} & 0.968(0.964)& \textbf{0.971(0.970)} & \textbf{0.974(0.971)} & \textbf{0.977(0.973)} \\
\hline
\end{tabular}
\end{adjustbox}
\begin{tablenotes}
\footnotesize
\item $^\dagger$ refers to significant differences (p\textless0.05) compared with the results of ACIR, * indicates the non-open source model reproduced by ourselves, and vice versa. Results are presented in mAP (maAP).
\end{tablenotes}
\vspace{-0.2cm}
\end{table}

\begin{figure}[htbp]
    \centering
    \includegraphics[width=\linewidth]{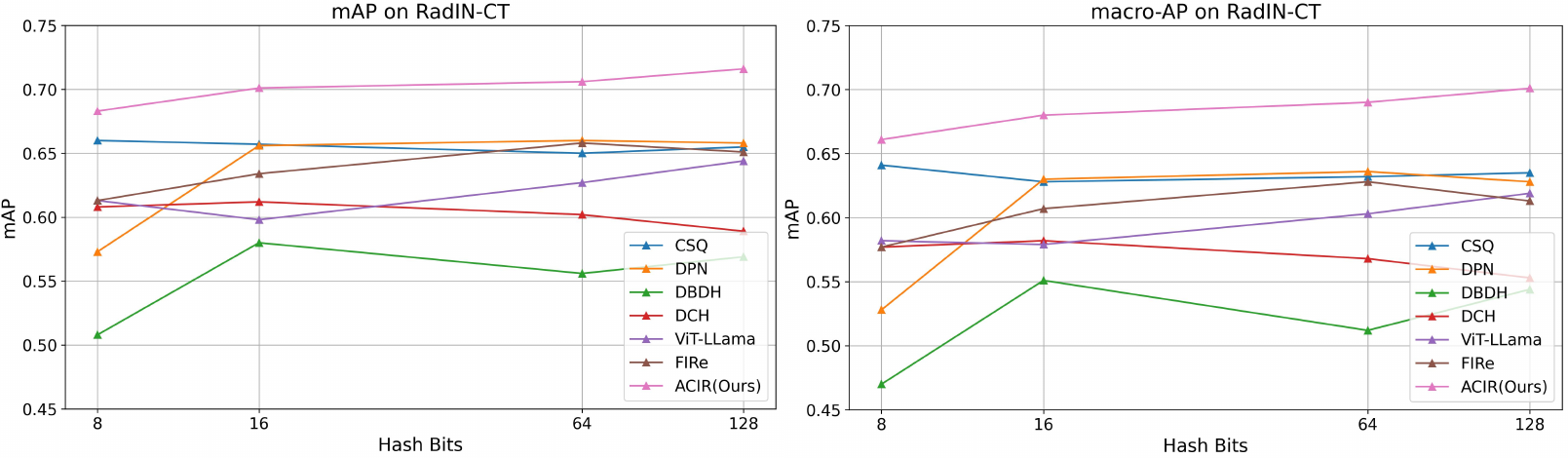}
    \vspace{-0.7cm}
    \caption{mean average precision (mAP) and macro average precision (maAP)on RadIN-CT dataset. All the models were trained according to their released codes.}
    \label{fig:enter-label}
    \vspace{-0.2cm}
\end{figure}

ACIR outperformed all comparison approaches, achieving the highest mAP of 0.716 and maAP of 0.701 in 128 bits, showing a significant improvement (p\textless0.05) of at least 5.6\% mAP compared to other approaches. CSQ achieved 0.660 mAP in 8 bits, outperforming its 16, 64 and 128-bit variants. FIRe gained 0.658 mAP and 0.628 maAP in 64 bits, followed by the vision language model ViT-LLaMa (achieving 0.644 mAP and 0.619 maAP in 128 bits). Interestingly, although QSMIH (0.492 mAP) and DPN (0.573) underperformed in compact bits, they achieved comparable results in 16 bits, 64 bits, and 128 bits. Unexpectedly, methods (ATH, MTH) developed for biomedical image retrieval exhibited low mAP and maAP, with 0.352 (0.316) and 0.560 (0.531) mAP(maAP), respectively. 

\noindent\textbf{Comparison experiments on BPS.}
The average performance of all models on the BPS dataset surpasses that of the RadIN-CT dataset, with several models showing notable achievements. The ViT-LLaMa model achieves high mAP scores across different bit levels, peaking at 0.971 with 128 bits. Most comparison approaches, excluding DBDH and ATH, demonstrated mAP exceeding 0.94. In addition to ACIR, ViT-LLaMa-128bit achieved the highest mAP (0.971) and maAP (0.969), followed by CSQ-8bit (0.97 mAP, 0.967 maAP) and MTH-8bit (0.968 mAP, 0.962 maAP). No significant difference (p \textgreater 0.05) was observed between ACIR-8bit and MTH-8bit on BPS, indicating a potential upper-bound retrieval capability with compact hash bits. Conversely, ATH and DBDH had relatively lower performance, with mAP peaking at 0.474 (64 bits) and 0.756 (8 bits), respectively.

\begin{table}[htbp]
\centering
\caption{Ablation Studies of ACIR}
\begin{tabular}{lccccccc}
\hline
CNN* & ViT* & Hybrid* & $L_{\mathrm{WSC}}$ & DaRF & KAN & CGR & mAP(maAP)\\
\hline
\checkmark &  & & & & & & 0.622 (0.602)  \\ 
\checkmark & & & \checkmark & & & & 0.653 (0.635)\\ 
 & \checkmark & & & & & & 0.595 (0.561) \\
 & \checkmark & & \checkmark & & & & 0.632 (0.594)\\
 &  & \checkmark & & & & & 0.645 (0.614)  \\
 &  & \checkmark & \checkmark & & & & 0.663 (0.643)  \\
 &  & \checkmark & \checkmark & \checkmark & & & 0.677 (0.658)  \\
 &  & \checkmark & \checkmark & \checkmark & \checkmark & & 0.686 (0.667)  \\
 &  & \checkmark & \checkmark & \checkmark & \checkmark & \checkmark  & 0.701 (0.680)  \\
\hline
\end{tabular}
\begin{tablenotes}
\footnotesize
\item * indicates using pretrained weights.
\end{tablenotes}
\label{tab:ablation}
\end{table}

\begin{figure*}[h]
    \centering
    \includegraphics[width=1\linewidth]{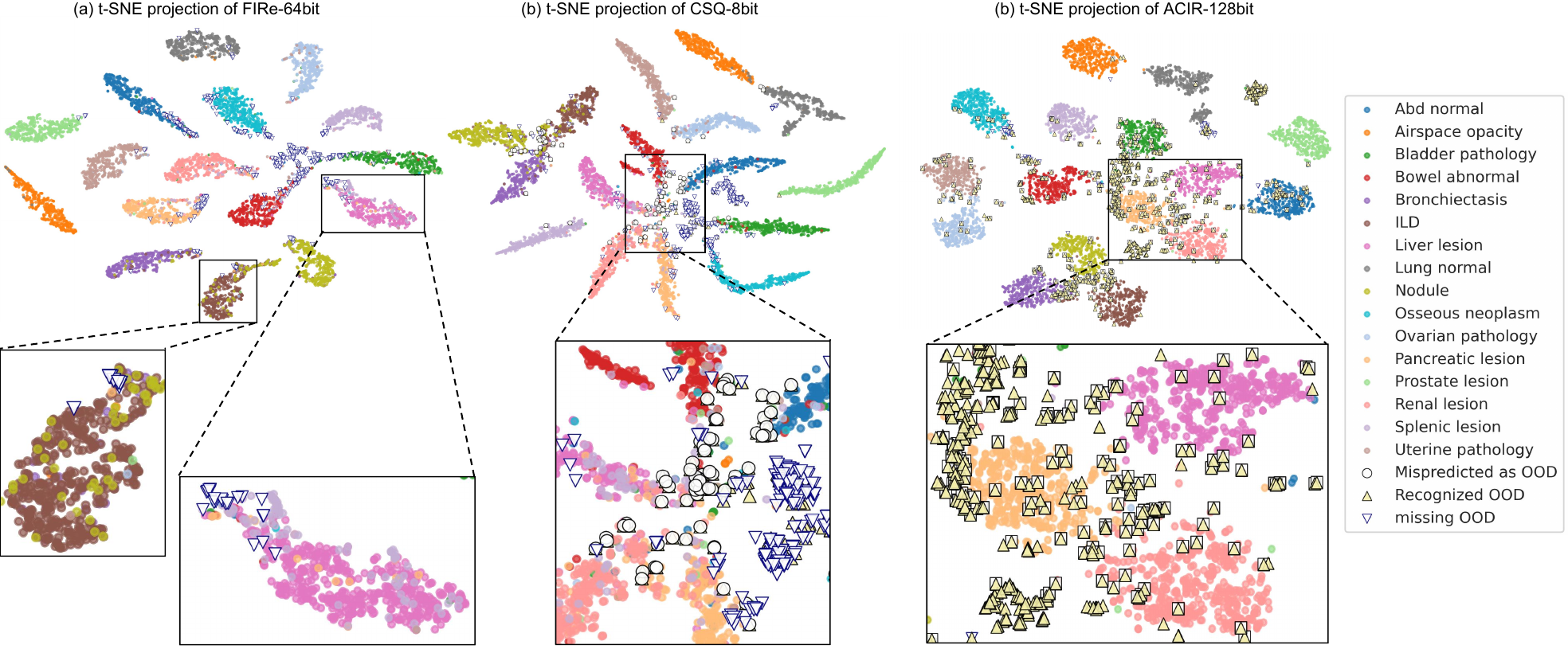}
    \caption{t-SNE projection of class distribution of (a) FIRe-64bit, (b) CSQ-8bit and (c) ACIR-128bit. The magnified windows highlight the distribution of the hash embeddings for renal, pancreatic, and liver lesions. ACIR presented much cleaner and tidier clusters compared with FIRe and CSQ, with fewer mispredictions and more recognized OOD samples. }
    \label{fig:tsne}
\end{figure*}

\noindent\textbf{Ablation studies.} Table \ref{tab:ablation} demonstrates the results of ablation experiments. The proposed hybrid model significantly outperforms (p\textless0.05) baseline models (CNN and ViT) with 0.645 mAP and 0.614 maAP, surpassing them by 2.3\% and 5\%, respectively. Models trained with a structure-aware contrastive loss $L_{WSC}$ consistently outperform their counterparts (p\textless0.05), especially for maAP. In particular, by applying $L_{WSC}$, ConvNet shows a 3.1\% (3.3\%) increase in mAP (maAP), ViT gains 3.8\% (3.3\%) in mAP (maAP), and the Hybrid model improves by 1.8\% (2.9\%). Notably, the Hybrid model was further improved by DaRF, with 1.4\% in mAP and 1.5\% in maAP. Additionally, instead of using MLP to aggregate hierarchical features, the introduction of the Kolmogorov-Arnold layer leads to an additional 0.9\% increase and 0.9\% in mAP and maAP, respectively. Notably, integrating content-guided ranking (CGR) resulted in a significant improvement in both mAP (1.5\% gain) and maAP (1.3\% gain). These results highlight the effectiveness of the method in enhancing the overall retrieval performance.

For hyper-parameter selection, both the mAP and maAP achieved the best performance when $\alpha=0.5$, with 0.701 and 0.680, respectively. In particular, the performance increased from $\alpha = 0$ (mAP: 0.662, maAP: 0.651) to $\alpha = 0.5$, reaching its peak, then decreasing when $\alpha = 0.75$ (mAP: 0.693, maAP: 0.672) and $\alpha = 1$ (mAP: 0.681, maAP: 0.658).

\noindent\textbf{Out-of-distribution assessment}
The results of the OOD evaluation (for the top-4 approaches) are presented in Table \ref{tab:ood_assess}. 
\begin{table}[htbp]
\vspace{-0.4cm}
\centering
\caption{Out-of-distribution assessment}
\begin{tabular}{lccc}
\hline
Models & Precision & Recall & OOD detection rate \\
\hline
DPN-64bit & 0.788 & 0.790 & 22/500 \\
CSQ-8bit & 0.849 & 0.860 & 80/500 \\
ViT-LLaMa-128bit & 0.839 & 0.848 & 85/500\\
FIRe-64bit & 0.870 & 0.856 & 2/500\\
ACIR-128bit w/o OOD & 0.869 & 0.880 & 87/500\\
ACIR-128bit & 0.908 & 0.901 & 402/500\\
\hline
\end{tabular}
\label{tab:ood_assess}
\vspace{-0.3cm}
\end{table}
We first calculated the hash centre for each category and assigned class labels based on predictions, followed by assessing the macro precision and recall. As Table \ref{tab:ood_assess} shows, ACIR-128bit achieved the best robustness with the highest precision of 90.2\%, recall of 0.901\%, and 80.4\% OOD detection rate (402 out of 500). Interestingly, although ViT-LLaMa-128bit undeformed FIRe-64bit in macro precision and recall, it recognized more OOD samples (85/500). 

\vspace{-0.2cm}
\section{Discussion}
In this study, we proposed a novel method that incorporates knowledge consolidation (using a hybrid ConvPiT and depth-aware fusion), structure-aware contrastive learning, and out-of-distribution (OOD) detection for medical image retrieval. The proposed method was thoroughly evaluated on pathological and radiological data, with a further evaluation on an OOD data set that includes natural, MR, cell, and retina images. Compared to existing studies, our method achieved significantly better results (p\textless 0.05) in addressing the challenges associated with medical image databases.

\noindent\textbf{Overall performance.} Results on the BPS dataset significantly outperformed those on the RadIN-CT dataset, indicating larger inter-class variation in BPS. The complex characteristic of RadIN-CT reflects the models' ability to accurately retrieve disease patterns among similar samples. This highlights the models' capacity to handle 'hard' datasets with more classes and fewer inter-class variations. For instance, MTH underperformed on RadIN-CT (9th place with 0.560 mAP) but showed strong results on BPS data (3rd place with 0.967 mAP), suggesting MTH struggles with complex datasets. Interestingly, CSQ performed competitively with compact hash codes, achieving competitive mAP (2nd place with 0.660) with 8 bits, but underperformed with 16, 64, and 128 bits. This indicates that longer predefined hash codes may result in performance reduction. ATH, following the official implementation, showed weak performance on both datasets for unknown reasons. In contrast, our proposed method demonstrated superior robustness and accuracy, achieving top performance across all bit levels and datasets. 

\begin{figure*}[!h]
    \centering
    \includegraphics[width=1\linewidth]{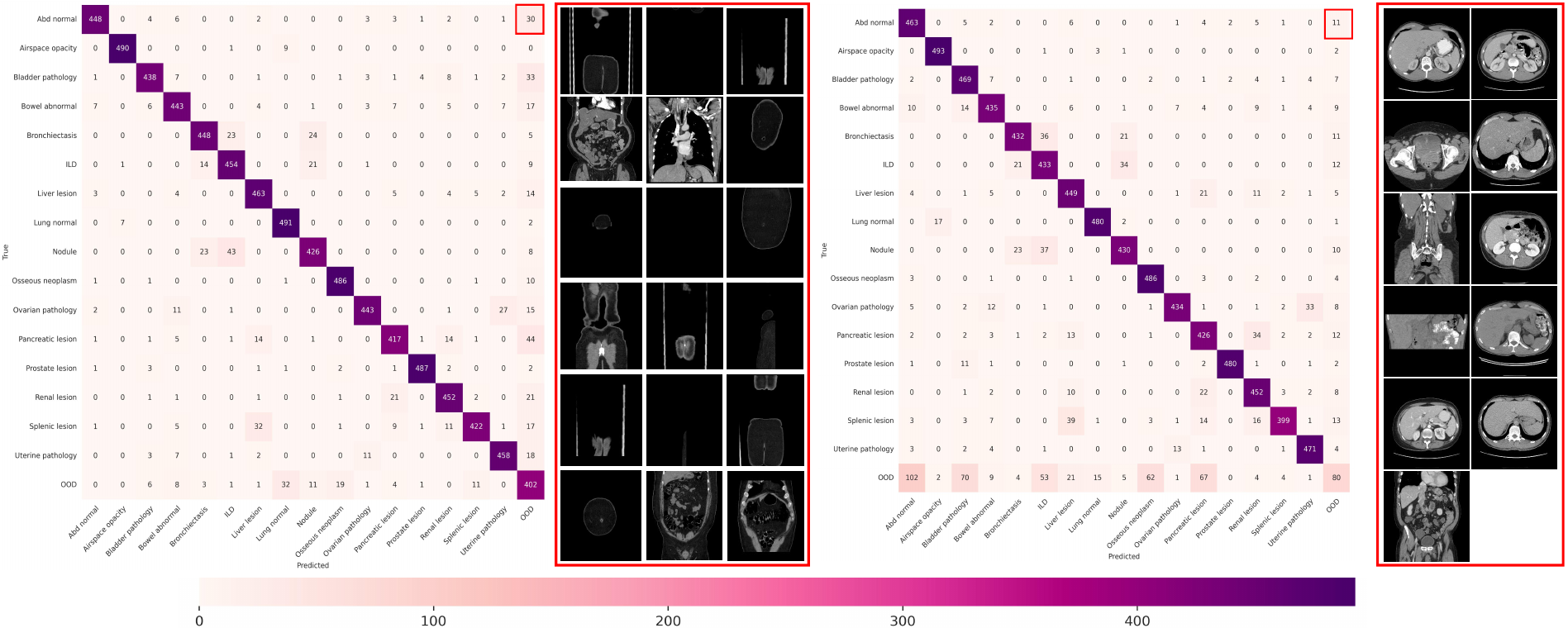}
    \caption{Confusion matrix of ACIR (left) and CSQ (right) with details of mispredicted samples on OOD dataset (classified as Abd normal but predicted as OOD). The false predictions suggest that ACIR has the capacity to identify low-quality or challenging samples.}
    \label{fig:oodscreen}
    \vspace{-0.4cm}
\end{figure*}

\noindent\textbf{Pretraining assessment.} In this study, Transformer/hybrid approaches (ACIR, ViT-LLaMa, and FIRe) have demonstrated comparable performance. 
\begin{table}[!h]
\vspace{-0.3cm}
\centering
\caption{Pretraining Assessment}
\begin{threeparttable}
\centering
\begin{tabular}{lcc}
\hline
Models & w/o Pretraining & Pretrained \\
\hline
CSQ-8bit & 0.618 (0.589) & 0.660 (0.641) \\
ViT-LLaMa-128bit & 0.304 (0.268) & 0.644 (0.619) \\
FIRe-64bit & 0.572 (0.545) & 0.658 (0.628) \\
ACIR-128bit & 0.627 (0.603)& 0.716 (0.701)\\
\hline
\end{tabular}
\begin{tablenotes}[flushleft]
\footnotesize
\item Results are presented in mAP (maAP).
\end{tablenotes}
\end{threeparttable}
\label{tab:pretraining}
\vspace{-0.3cm}
\end{table}
In addition to the SOTA ACIR, ViT-LLaMa ranked 5th (0.644 mAP) on RadIN-CT and 2nd (0.971 mAP) on BPS. Similarly, FIRe secured 4th place (0.658 mAP) on RadIN-CT and 6th place (0.958 mAP) on BPS. Since most Transformers require transfer learning, a question arises: will they maintain good performance when training from scratch? To address this, we further investigated the performance of Transformer and hybrid models without pretraining, as detailed in Table \ref{tab:pretraining}.

Interestingly, without pretraining, Transformers and hybrid models underperform ConvNets. Specifically, ViT-LLaMa and hybrid models (FIRe and ACIR) saw performance drops of nearly 34\% and 9\%, respectively, while ConvNets only dropped by around 5\%. This indicates that Transformers and hybrid models rely more heavily on vast data or pretrained weights. Without pretraining, these models may get trapped in local optima and exhibit slow convergence.

\noindent\textbf{Knowledge consolidation.} The results in Table \ref{tab:ablation} suggest the importance of knowledge consolidation in representation learning for retrieval models. By consolidating globally refined features (Hybrid*) and multi-level representations (DaRF), the retrieval performance witnessed a significant improvement with 2.3\% and 6\% mAP compared with CNN and ViT, respectively. 

\noindent\textbf{Out-of-distribution performance.} We used t-SNE \cite{van2008visualizing} to visualize the distribution of hash embeddings from top-performing models (FIRe, CSQ, and ACIR). Features for t-SNE were derived from the representations before the last layer (KAN for ACIR, Linear for FIRe and CSQ). ACIR demonstrated strong robustness and achieved state-of-the-art performance. Its class clusters were tidier and cleaner, with most OOD samples successfully recognized (Fig \ref{fig:tsne}). In contrast, the OOD samples for FIRe and CSQ were closely clustered around the known classes, indicating weak capabilities for models without knowledge consolidation of extracting distinguishable features against OOD samples. Moreover, the class distributions of FIRe and CSQ were more dispersed with elongated shapes. In contrast, the class distribution of ACIR shows more compact clustering, suggesting that the consolidated features are more distinguishable than the comparison ones. Additionally, features of different classes were intermixed (see magnified windows) for FIRe and CSQ, illustrating poor capacity against OOD samples and lower accuracy in feature representation.

\begin{figure*}[htbp]
    \centering
    \includegraphics[width=1\linewidth]{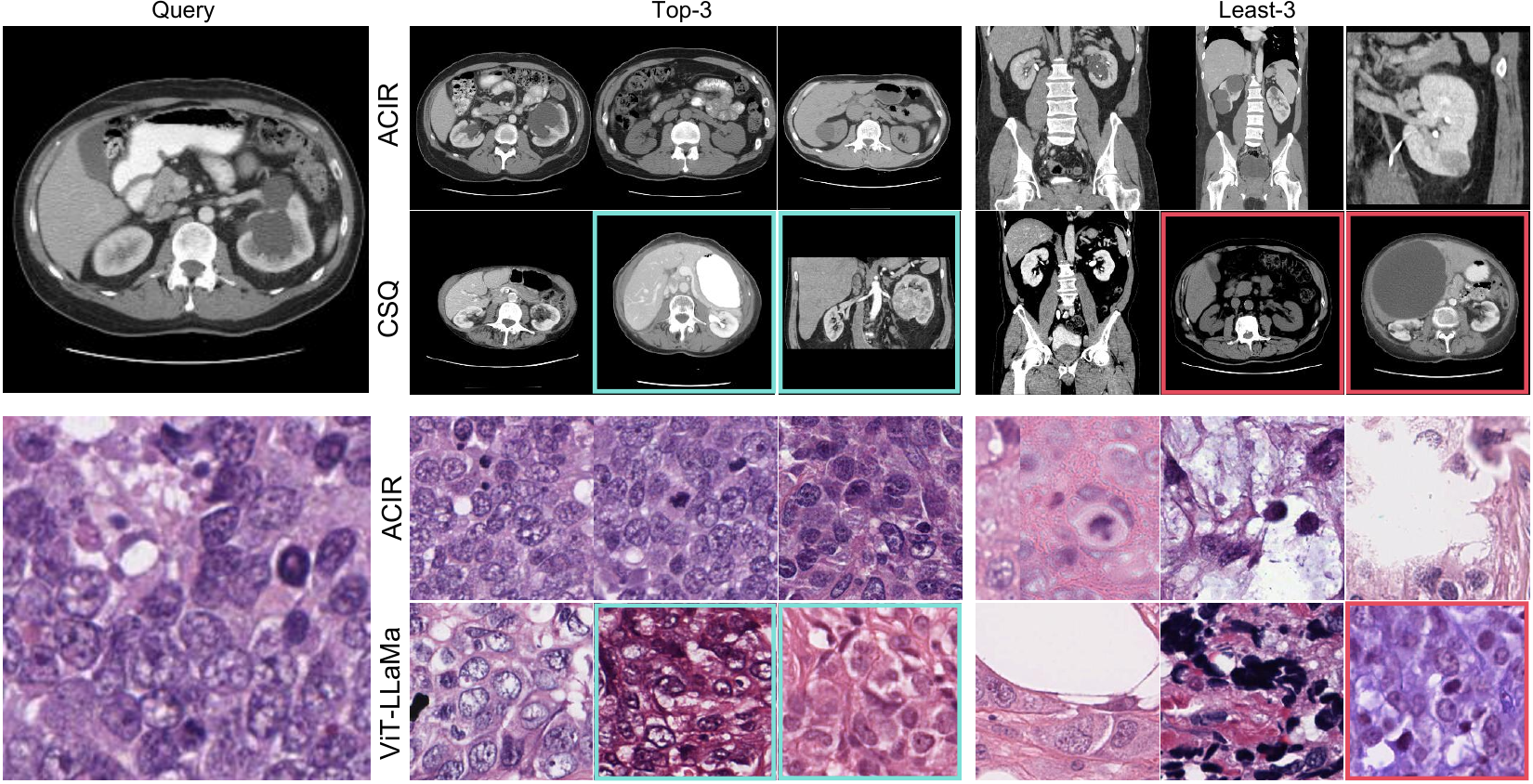}
    \caption{Ranked retrieval results given by ACIR, CSQ (2nd place on RadIN-CT), and ViT-LLaMa (2nd place on BPS). The top-3 similar and dissimilar samples for the query images were presented. Samples with cyan boxes indicate the top-3 similar samples (to the query) while with considerable visual differences exist. Samples with red boxes indicate the least-3 retrieved samples with certain visual similarities.}
    \label{fig:rank}
    \vspace{-0.4cm}
\end{figure*}

Despite ACIR's strong ability to identify OOD samples, it mistakenly categorises some samples as OOD. Interestingly, the majority of these samples were either mislabeled, of poor quality, or challenging with confusing patterns or traits (Fig. \ref{fig:oodscreen}), indicating ACIR's resilience against anomalies. In comparison, erroneous predictions made by CSQ are shown in Fig. \ref{fig:oodscreen} (right), attributed to its similar performance in recognising OOD samples (Table \ref{tab:ood_assess}). Contrary to ACIR, the OOD samples incorrectly predicted by CSQ lacked specific characteristics, and no low-quality samples were detected.

\noindent\textbf{Upper bound performance.} While certain models achieved better performance with an increase in the number of hash bits, excessively long hash bits could lead to a decline in performance.
\begin{table}[!h]
\vspace{-0.3cm}
\centering
\caption{Sparse hash bits assessment}
\begin{threeparttable}
\centering
\begin{tabular}{p{2.5cm}cc}
\hline
Models & 256 bit & 512 bit \\
\hline
QSMIH & 0.689 (0.652) & 0.702 (0.673) \\
ViT-LLaMa & 0.661 (0.634) & 0.679 (0.654) \\
ACIR & 0.724 (0.707) & 0.723 (0.701)\\
\hline
\end{tabular}
\begin{tablenotes}[flushleft]
\footnotesize
\item Results are presented in mAP (maAP).
\end{tablenotes}
\end{threeparttable}
\label{tab: sparse assess}
\vspace{-0.5cm}
\end{table}
Particularly, models such as DBDH, DCH, DSDH, and ATH displayed superior performance at 16 bits compared to at 8 bits. Alternatively, models like DPN, FIRe, and MTH reached their optimal performance at 64 bits but then showed a notable decrease in accuracy when the hash length increased to 128 bits. Moreover, QSMIH, ViT-LLaMa, and the newly proposed ACIR demonstrated potential for further enhancements. Consequently, we carried out additional experiments on these three models to determine their maximum capabilities. The findings were interesting, with all three models exhibiting significant improvements at 256 bits, while the trends at 512 bits were various. This indicates that the sparse hash codes generated through these methodologies can retain more detailed information. Furthermore, this opens a new potential research question: 'Considering the rapid progress in hardware and ongoing updates in storage technology, is it still necessary to compress data into compact hash codes?'

\noindent\textbf{Ranked retrieval.}
The top-3 similar and dissimilar samples were selected from the retrieved results, and sorted by Hamming distance between the hash codes of the database samples and the query image. We evaluated three approaches: ACIR, CSQ (the 2nd place model on RadIN-CT), and ViT-LLaMa (the 2nd place model on BPS), to explore the ranked retrieval performance of different methods.

For each comparison, the top-3 similar and dissimilar samples for the query images were presented (in Fig. \ref{fig:rank}). ACIR's retrieval results are well given based on similarity, while those of CSQ and ViT-LLaMa are confused. For instance, with a query image depicting renal pathology, ACIR efficiently identifies images with similar patterns. In contrast, both CSQ and ViT-LLaMa failed to sort the retrieved results accurately based on image contents. Additionally, the most similar samples retrieved by these two models exhibited significant visual differences compared to the query image.

To further explore the capabilities of different models against the over-centralized issue, we calculated the accuracy, precision and recall of top-4 approaches under different decision boundaries (different radii of hamming balls). Ideally, a robust model capable of addressing the over-centralization issue should demonstrate a simultaneous increase in precision and recall as the radius R of the Hamming ball (the decision boundary) expands. Alternatively, it may exhibit stable or slightly reduced precision accompanied by an increase in recall. In contrast, models suffering from over-centralization tend to concentrate most samples near the cluster centre. As a result, increasing the decision boundary R will have a limited impact on recall, since the majority of samples are already within the core region.

From Table 6, we observe that CSQ performs notably well when R=0. However, once the decision boundary R (radius) increases from 0 to 1, CSQ’s precision and recall only show modest improvements—revealing a tendency toward over-centralization. Specifically, CSQ pushes samples within the same class too tightly around a single centroid; thus, when the decision boundary is relaxed (from R=0 to R=1), CSQ fails to expand its coverage of reasonably dispersed samples. 
\begin{table}[h]
    \centering
\caption{Model performance of different comparisons.}
\label{tab:model_performance}
\begin{adjustbox}{width=\textwidth}
    \begin{tabular}{ccccc}
    \hline
    & \textbf{CSQ-16bit} & \textbf{ViT-LLaMa-16bit} & \textbf{FIRe-16bit} & \textbf{ACIR-16bit} \\ 
    \hline
    R=0 & 0.715 (0.820/0.715) & 0.377 (0.685/0.384) & 0.515 (0.820/0.517) & 0.555 (0.832/0.532) \\ 
    R=1 & 0.733 (0.791/0.733) & 0.566 (0.719/0.570) & 0.640 (0.774/0.631) & 0.676 (0.810/0.673) \\ 
    R=2 & 0.742 (0.784/0.743) & 0.626 (0.706/0.624) & 0.676 (0.747/0.663) & 0.723 (0.807/0.719) \\ 
    R=3 & 0.750 (0.775/0.750) & 0.675 (0.698/0.669) & 0.689 (0.733/0.675) & 0.753 (0.794/0.742) \\ 
    R=4 & 0.757 (0.767/0.757) & 0.685 (0.690/0.679) & 0.692 (0.731/0.678) & 0.766 (0.780/0.766) \\ 
    R=5 & 0.757 (0.767/0.757) & 0.689 (0.688/0.684) & 0.692 (0.731/0.678) & 0.770 (0.770/0.765) \\ 
    R=6 & 0.757 (0.767/0.757) & 0.689 (0.688/0.684) & 0.692 (0.731/0.678) & 0.770 (0.770/0.765) \\ 
    R=7 & 0.757 (0.767/0.757) & 0.689 (0.688/0.684) & 0.692 (0.731/0.678) & 0.770 (0.770/0.766) \\ 
    \hline
    \end{tabular}
    \end{adjustbox}
    \begin{tablenotes}[flushleft]
\footnotesize
\item  Results are shown in acc (prec/rec) scores.
\end{tablenotes}
\end{table}

In contrast, ViT-LLama, FIRe, and ACIR demonstrate substantial increases in recall when RRR moves from 0 to 1, indicating they are more robust against the over-centralization issue. Among these, ACIR stands out with consistently higher accuracy and recall across a broad range of radii. This suggests that ACIR’s modelling approach better reflects the actual distribution of samples. Since the “class centre” is defined by averaging all hash codes belonging to a class, relying too heavily on a tight centroid (as CSQ does) may yield short-term gains at R=0 but neglects the natural dispersion of samples within a class. By contrast, ACIR maintains high performance even as the radius grows, incorporating more samples that correctly belong to each class. As a result, ACIR not only achieves strong accuracy and recall but also demonstrates the capability against over-centralization.

\noindent\textbf{Computational cost and inference efficiency.} Besides the accuracy, one major concern for deep learning models is the computational cost and inference efficiency. Here we report FLOPs (Floating Point Operations), \#Param (the number of trainable parameters), and inference time for top-4 approaches (including CSQ, ViT-LLaMa, FIRe and ACIR), shown in Table VII (bit configurations only influence the last layer of the model, therefore the computational cost across these models is similar  ).
\begin{table}[!h]
\vspace{-0.3cm}
\centering
\caption{Efficiency and scale assessment}
\begin{threeparttable}
\centering
\begin{tabular}{lcccc}
\hline
Attributes & CSQ & ViT-LLaMa & FIRe & ACIR \\
\hline
FLOPs (GFLOPs) & 4.13  & 44.7 & 7.87 & 12.44 \\
\#Param (M) & 25.56 & 227.11 & 105.44 & 151.33 \\
Inference time (s) & 0.027  & 0.085 & 0.055 & 0.080 \\
\hline
\end{tabular}
\end{threeparttable}
\label{tab: sparse assess}
\vspace{-0.3cm}
\end{table}
The results indicate that although the proposed ACIR achieved state-of-the-art performance, it is less efficient than CSQ and FIRe. However, the superior performance of ACIR suggests that despite its relative complexity, ACIR offers greater performance potential and a higher upper limit. This is precisely what is needed for clinical applications. Furthermore, ViT-LLaMa achieved fourth place in both efficiency and performance, indicating its relatively weak capacity compared to other top solutions. 

\noindent\textbf{Image retrieval and data harmonisation.} Data harmonisation refers to removing the non-biological bias of multicentre datasets \cite{nan2022data, schmidt2020definitions}, while image retrieval plays an essential role as a preliminary step. Effective data harmonisation is crucial in realizing the full potential of AI models, particularly in multicentre studies where data inconsistency can significantly hinder model performance. Ideally, data with similar contents/targets should be harmonized together to ensure samples of the same category are allocated in the same distribution. For instance, in some interdisciplinary fields, researchers have used synthetic models to align cross-modal data to eliminate spatial-spectral discrepancies \cite{sun2021multisensor}. With content-based retrieval, content-based harmonisation could be achieved to identify and adjust for these variances, ensuring downstream tasks are more accurate and reliable. By improving how models handle diverse datasets, we can facilitate broader collaboration in the medical community and accelerate the pace of discovery and innovation. 

\textbf{Limitations and future research directions.} In image retrieval, evaluation typically relies on calculating the metrics such as the mAP. These metrics are well-suited for scenarios where retrieval relevance can be directly assessed using class labels. However, in the context of content-based image recommendation, class labels alone are insufficient to determine content similarity, as the recommendation task emphasizes perceptual or structural resemblance rather than categorical matches. Consequently, there is a need for a novel evaluation metric that not only measures retrieval accuracy but also provides an assessment of content similarity. This metric should incorporate factors such as visual similarity, semantic consistency, and structural coherence to reflect the nuanced nature of image recommendations. Developing such an evaluation framework will enable a more comprehensive understanding of the system's ability to identify content-relevant images, bridging the gap between traditional retrieval metrics and the requirements of content-based recommendation systems. In addition to supervised approaches which require further training procedures, unsupervised methods offer a promising alternative, particularly in scenarios where annotated data is limited or unavailable \cite{sun2022unsupervised, nan2022unsupervised}. However, challenges persist regarding the stability and reproducibility of unsupervised algorithms, as well as their ability to consistently bridge the gap between perceptual similarity and semantic content.

\vspace{-0.3cm}
\section{Conclusion}
This paper introduces an innovative and resilient image recommendation system that produces ordered retrieval outcomes based on content similarity. The suggested ACIR creatively consolidates hierarchical knowledge into blended representations through a Depth-aware Representation Fusion and Structure-aware Contrastive Hashing. The OOD detection model has significantly improved the model's retrieval capacity, especially for out-of-distribution query samples. Extensive experiments have demonstrated that ACIR achieved state-of-the-art performance in context-based retrieval and OOD recognition on both pathology and radiology datasets. The results suggest that ACIR is a promising and adaptable solution for practical image retrieval applications in various imaging environments. 

\section{Acknowledgements}
This work was supported in part by the ERC IMI (101005122), the H2020 (952172), the MRC (MC/PC/21013), the Royal Society (IEC\textbackslash NSFC\textbackslash 211235), the NVIDIA Academic Hardware Grant Program, the SABER project supported by Boehringer Ingelheim Ltd, NIHR Imperial Biomedical Research Centre (RDA01), Wellcome Leap Dynamic Resilience, UKRI guarantee funding for Horizon Europe MSCA Postdoctoral Fellowships (EP/Z002206/1), and the UKRI Future Leaders Fellowship (MR/V023799/1).
.

\bibliographystyle{elsarticle-num} 
\bibliography{ref}
\end{document}